# A Stochastic Decoder for Neural Machine Translation[*]

**Philip Schulz**
Amazon Research[†]
phschulz@amazon.com

**Wilker Aziz**
University of Amsterdam
w.aziz@uva.nl

**Trevor Cohn**
University of Melbourne
trevor.cohn@unimelb.edu.au

## Abstract

The process of translation is ambiguous, in that there are typically many valid translations for a given sentence. This gives rise to significant variation in parallel corpora, however, most current models of machine translation do not account for this variation, instead treating the problem as a deterministic process. To this end, we present a deep generative model of machine translation which incorporates a chain of latent variables, in order to account for local lexical and syntactic variation in parallel corpora. We provide an in-depth analysis of the pitfalls encountered in variational inference for training deep generative models. Experiments on several different language pairs demonstrate that the model consistently improves over strong baselines.

## 1 Introduction

Neural architectures have taken the field of machine translation by storm and are in the process of replacing phrase-based systems. Based on the encoder-decoder framework (Sutskever et al., 2014) increasingly complex neural systems are being developed at the moment. These systems find new ways of extracting information from the source sentence and the target sentence prefix for example by using convolutions (Gehring et al., 2017) or stacked self-attention layers (Vaswani et al., 2017). These architectural changes have led to great performance improvements over classical RNN-based neural translation systems (Bahdanau et al., 2014).

Surprisingly, there have been almost no efforts to change the probabilistic model wich is used to train the neural architectures. A notable exception is the work of Zhang et al. (2016) who introduce a sentence-level latent Gaussian variable.

In this work, we propose a more expressive latent variable model that extends the attention-based architecture of Bahdanau et al. (2014). Our model is motivated by the following observation: translations by professional translators vary across translators but also within a single translator (the same translator may produce different translations on different days, depending on his state of health, concentration etc.). Neural machine translation (NMT) models are incapable of capturing this variation, however. This is because their likelihood function incorporates the statistical assumption that there is one (and only one) output[1] for a given source sentence, i.e.,

$$P(y_1^n|x_1^m) = \prod_{i=1}^n P(y_i|x_1^m, y_{<i}) \,. \quad (1)$$

Our proposal is to augment this model with latent sources of variation that are able to represent more of the variation present in the training data. The noise sources are modelled as Gaussian random variables.

The contributions of this work are:

- The introduction of an NMT system that is capable of capturing word-level variation in translation data.
- A thorough discussions of issues encountered when training this model. In particular, we motivate the use of KL scaling as introduced by Bowman et al. (2016) theoretically.

---

[*] Code and a workflow that reproduces the experiments are available at https://github.com/philschulz/stochastic-decoder.

[†] Work done prior to joining Amazon.

[1] Notice that from a statistical perspective the output of an NMT system is a distribution over target sentences and not any particular sentence. The mapping from the output distribution to a sentence is performed by a decision rule (e.g. argmax decoding) which can be chosen independently of the NMT system.

- An empirical demonstration of the improvements achievable with the proposed model.

## 2 Neural Machine Translation

The NMT system upon which we base our experiments is based on the work of Bahdanau et al. (2014). The likelihood of the model is given in Equation (1). We briefly describe its architecture.

Let $x_1^m = (x_1, \ldots, x_m)$ be the source sentence and $y_1^n$ the target sentence. Let $\text{RNN}(\cdot)$ be any function computed by a recurrent neural network (we use a bi-LSTM for the encoder and an LSTM for the decoder). We call the decoder state at the $i$th target position $t_i$; $1 \leq i \leq n$. The computation performed by the baseline system is summarised below.

$$[h_1, \ldots, h_m] = \text{RNN}(x_1^m) \tag{2a}$$
$$\tilde{t}_i = \text{RNN}(t_{i-1}, y_{i-1}) \tag{2b}$$
$$e_{ij} = v_a^\top \tanh\left(W_a[\tilde{t}_i, h_j]^\top + b_a\right) \tag{2c}$$
$$\alpha_{ij} = \frac{\exp(e_{ij})}{\sum_{j=1}^m \exp(e_{ij})} \tag{2d}$$
$$c_i = \sum_{j=1}^m \alpha_{ij} h_j \tag{2e}$$
$$t_i = W_t[\tilde{t}_i, c_i]^\top + b_t \tag{2f}$$
$$\phi_i = \text{softmax}(W_o t_i + b_o) \tag{2g}$$

The parameters $\{W_a, W_t, W_o, b_a, b_t, b_o, v_a\} \subseteq \theta$ are learned during training. The model is trained using maximum likelihood estimation. This means that we employ a cross-entropy loss whose input is the probability vector returned by the softmax.

## 3 Stochastic Decoder

This section introduces our stochastic decoder model for capturing word-level variation in translation data.

### 3.1 Motivation

Imagine an idealised translator whose translations are always perfectly accurate and fluent. If an MT system was provided with training data from such a translator, it would still encounter variation in that data. After all, there are several perfectly accurate and fluent translations for each source sentence. These can be highly different in both their lexical as well as their syntactic realisations.

In practice, of course, human translators' performance varies according to their level of education, their experience on the job, their familiarity with the textual domain and myriads of other factors. Even within a single translator variation may occur due to level of stress, tiredness or status of health. That translation corpora contain variation is acknowledged by the machine translation community in the design of their evaluation metrics which are geared towards comparing one machine-generated translation against several human translations (see e.g. Papineni et al., 2002).

Prior to our work, the only attempt at modelling the latent variation underlying these different translations was made by Zhang et al. (2016) who introduced a sentence level Gaussian variable. Intuitively, however, there is more to latent variation than a unimodal density can capture, for example, there may be several highly likely clusters of plausible variations. A cluster may e.g. consist of identical syntactic structures that differ in word choice, another may consist of different syntactic constructs such as active or passive constructions. Multimodal modelling of these variations is thus called for—and our results confirm this intuition.

An example of variation comes from free word order and agreement phenomena in morphologically rich languages. An English sentence with rigid word order may be translated into several orderings in German. However, all orderings need to respect the agreement relationship between the main verb and the subject (indicated by underlining) as well as the dative case of the direct object (dashes) and the accusative of the indirect object (dots). The agreement requirements are fixed and independent of word order.

1. I can't imagine you naked.
   (a) Ich kann mir dich nicht nackt vorstellen.
   (b) Ich kann dich mir nicht nackt vorstellen.
   (c) Dich kann ich mir nicht nackt vorstellen.

Stochastically encoding the word order variation allows the model to learn the same agreement phenomenon from different translation variants as it does not need to encode the word order and agreement relationships jointly in the decoder state.

Further examples of VP and NP variation from an actual translation corpus are shown in Figure 1.

We aim to address these word-level variation phenomena with a stochastic decoder model.

| |  |
|---|---|
| 预计听证会将进行两天。 | VOM19981105_0700_0262 |

The hearing is expected to last two days.
The hearing will last two days.
The hearings are expected to last two days.
It is expected that the hearing will go on for two days.

| |  |
|---|---|
| 众议院共和党的起诉人则希望传唤莱温斯基等多达15个人出庭作证。 | VOM19981230_0700_0515 |

However, the Republican complainant in the House wanted to summon 15 people including Lewinsky to testify in court.
The prosecutor of Republican Party in House of Representative hoped to summons more than 15 persons, including Lewinsky, to court.
The House of Representatives republican prosecution hopes to summon over fifteen witnesses including Monica Lewinsky to appear in court.

Figure 1: Examples from the multiple-translation Chinese corpus (LDC2002T01), where the translations come from different translators. These demonstrate the lexical variation of the verb and variation between passive and raising structures (top), and lexical variation on the agent NP (bottom). Both examples also exhibit appreciable length variation.

### 3.2 Model formulation

The model contains a latent Gaussian variable for each target position. This variable depends on the previous latent states and the decoder state. Through the use of recurrent networks, the conditioning context does not need to be restricted and the likelihood factorises exactly.

$$P(y_1^n|x_1^m) = \int \mathrm{d}z_0^n \, p(z_0|x_1^m) \times \\ \prod_{i=1}^n p(z_i|z_{<i}, y_{<i}, x_1^m) P(y_i|z_1^i, y_{<i}, x_1^m) \quad (3)$$

As can be seen from Equation (3), the model also contains a 0th latent variable that is meant to initialise the chain of latent variables based solely on the source sentence. Contrast this with the model of Zhang et al. (2016) which uses *only* that 0th variable.

A graphical representation of the stochastic decoder model is given in Figure 2a. Its generative story is as follows

$$Z_0|x_1^m \sim \mathcal{N}(\mu_0, \sigma_0^2) \quad (4a)$$
$$Z_i|z_{<i}, y_{<i}, x_1^m \sim \mathcal{N}(\mu_i, \sigma_i^2) \quad (4b)$$
$$Y_i|z_0^i, y_{<i}, x_1^m \sim \mathrm{Cat}(\phi_i) \quad (4c)$$

where $i = 1, \ldots, n$ and both the Gaussian and the Categorical parameters are predicted by neural network architectures whose inputs vary per time step. This probabilistic formulation can be implemented with a multitude of different architectures. We present ours in the next section.

### 3.3 Neural Architecture

Since the model contains latent variables and is parametrised by a neural network, it falls into the class of deep generative models (DGMs). We use a reparametrisation of the Gaussian variables (Kingma and Welling, 2014; Rezende et al., 2014; Titsias and Lázaro-Gredilla, 2014) to enable back-propagation inside a stochastic computation graph (Schulman et al., 2015). In order to sample $d$-dimensional Gaussian variable $z \in \mathbb{R}^d$ with mean $\mu$ and variance $\sigma^2$, we first sample from a standard Gaussian distribution and then transform the sample,

$$z = \mu + \sigma \odot \epsilon \qquad \epsilon \sim \mathcal{N}(0, \mathrm{I}) \ . \quad (5)$$

Here $\mu, \sigma \in \mathbb{R}^d$ and $\odot$ denotes element-wise multiplication (also known as Hadamard product). See the supplement for details on the Gaussian reparametrisation.

We use neural networks with one hidden layer with a tanh activation to compute the mean and standard deviation of each Gaussian distribution. A softplus transformation is applied to the output of the standard deviation's network to ensure positivity. Let us denote the functions that these networks compute by $f$.

For the initial latent state $z_0$ we compute the mean and standard deviation as

$$\mu_0 = f_{\mu_0}(h_m) \qquad \sigma_0 = f_{\sigma_0}(h_m) \ . \quad (6)$$

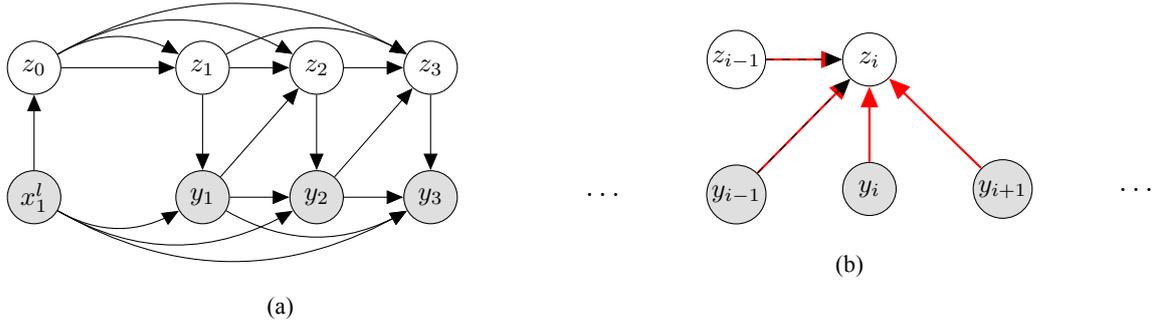

Figure 2: Graphical representation of 2a the generative model and 2b the inference model. Black lines indicate generative parameters ($\theta$) and red lines variational parameters ($\lambda$). Dashed red-black lines indicate that the inference model uses feature representations computed by the generative model as inputs. Through the recurrent net, the generative model (2a) also conditions its outputs on all previous latent assignments. We omit these arrows to avoid clutter. The inference model (2b) is only used at training time. Dots indicate further conditioning context.

The parameters of all other latent distributions are computed by functions $f_\mu$ and $f_\sigma$ whose inputs vary per target position.

$$\mu_i = f_\mu(t_{i-1}, z_{i-1}) \quad \sigma_i = f_\sigma(t_{i-1}, z_{i-1}) \quad (7)$$

Using these values, each latent variable is sampled according to Equation (5). The sampled latent variables are then used to modify the update of the decoder hidden state (Equation (2b)) as follows:

$$\tilde{t}_i = \text{RNN}(t_{i-1}, y_{i-1}, z_i) \quad (8)$$

The remaining computations stay unchanged. Notice that the latent values are used directly in updating the decoder state. This makes the decoder state a function of a random variable and thus the decoder state is itself random. Applying this argument recursively shows that also the attention mechanism is random, making the decoder entirely stochastic.

## 4 Inference and Training

We use variational inference (see e.g. Blei et al., 2017) to train the model. In variational inference, we employ a variational distribution $q(z)$ that approximates the true posterior $p(z|x)$ over the latent variables. The distribution $q(z)$ has its own set of parameters $\lambda$ that is disjoint from the set of model parameters $\theta$. It is used to maximise the evidence lower bound (ELBO) which is a lower bound on the marginal likelihood $p(x)$. The ELBO is maximised with respect to both the model parameters $\theta$ and the variational parameters $\lambda$.

Most NLP models that use DGMs only use one latent variable (e.g. Bowman et al., 2016). Models that use several variables usually employ a mean field approximation under which all latent variables are independent. This turns the ELBO into a sum of expectations (e.g. Zhou and Neubig, 2017). For our stochastic decoder we design a more flexible approximation posterior family which respects the dependencies between the latent variables,

$$q(z_0^n) = q(z_0) \prod_{i=1}^{n} q(z_i | z_{<i}) . \quad (9)$$

Our stochastic decoder can be viewed as a stack of conditional DGMs (Sohn et al., 2015) in which the latent variables depend on one another. The ELBO thus consists of nested positional ELBOs,

$$\begin{aligned} \text{ELBO}_0 + \mathbb{E}_{q(z_0)}[\text{ELBO}_1 \\ + \mathbb{E}_{q(z_1)}[\text{ELBO}_2 + \ldots]] , \end{aligned} \quad (10)$$

where for a given target position $i$ the ELBO is

$$\begin{aligned} \text{ELBO}_i = \mathbb{E}_{q(z_i)} [\log p(y_i | x_1^m, y_{<i}, z_{<i}, z_i)] \\ - \text{KL}\left(q(z_i) \,||\, p(z_i | x_1^m, y_{<i}, z_{<i})\right) . \end{aligned} \quad (11)$$

The first term is often called *reconstruction* or *likelihood* term whereas the second term is called the *KL* term. Since the KL term is a function of two Gaussian distributions, and the Gaussian is an exponential family, we can compute it analytically (Michalowicz et al., 2014), without the need for sampling. This is very similar to the hierarchical latent variable model of Rezende et al. (2014).

Following common practice in DGM research, we employ a neural network to compute the variational distributions. To discriminate it from the generative model, we call this neural net the *inference model*. At training time both the source and target sentence are observed. We exploit this by endowing our inference model with a "look-ahead" mechanism. Concretely, samples from the inference network condition on the information available to the generation network (Section 3.3) and also on the target words that are yet to be processed by the generative decoder. This allows the latent distribution to not only encode information about the currently modelled word but also about the target words that follow it. The conditioning of the inference network is illustrated graphically in Figure 2b.

The inference network produces additional representations of the target sentence. One representation encodes the target sentence bidirectionally (12a), in analogy to the source sentence encoding. The second representation is built by encoding the target sentence in reverse (12b). This reverse encoding can be used to provide information about future context to the decoder. We use the symbols $b$ and $r$ for the bidirectional and reverse target encodings, respectively. In our experiments, we again use LSTMs to compute these encodings.

$$[b_1, \ldots, b_n] = \text{RNN}(y_1^n) \quad (12a)$$
$$[r_1, \ldots, r_n] = \text{RNN}(y_1^n) \quad (12b)$$

In analogy to the generative model (Section 3.3), the inference network uses single hidden layer networks to compute the mean and standard deviations of the latent variable distributions. We denote these functions $g$ and again employ different functions for the initial latent state and all other latent states.

$$\mu_0 = g_{\mu_0}(h_m, b_n) \quad (13a)$$
$$\sigma_0 = g_{\sigma_0}(h_m, b_n) \quad (13b)$$
$$\mu_i = g_\mu(t_{i-1}, z_{i-1}, r_i, y_i) \quad (13c)$$
$$\sigma_i = g_\sigma(t_{i-1}, z_{i-1}, r_i, y_i) \quad (13d)$$

As before, we use Equation (5) to sample from the variational distribution.

During training, all samples are obtained from the inference network. Only at test time do we sample from the generator. Notice that since the inference network conditions on representations produced by the generator network, a naïve application of backpropagation would update parts of the generator network with gradients computed for the inference network. We prevent this by blocking gradient flow from the inference net into the generator.

### 4.1 Analysis of the Training Procedure

The training procedure as outlined above does not work well empirically. This is because our model uses a **strong generator**. By this we mean that the generation model (that is the baseline NMT model) is a very good density model in and by itself and does not need to rely on latent information to achieve acceptable likelihood values during training. DGMs with strong generators have a tendency to not make use of latent information (Bowman et al., 2016). This problem went initially unnoticed because early DGMs (Kingma and Welling, 2014; Rezende et al., 2014) used **weak generators**[2], i.e. models that made very strong independence assumptions and were not able to capture contextual information without making use of the information encoded by the latent variable.

Why DGMs would ignore the latent information can be understood by considering the KL-term of the ELBO. In order for the latent variable to be informative about the observed data, we need them to have high mutual information $I(Z; Y)$.

$$I(Z; Y) = \mathbb{E}_{p(z,y)} \left[ \log \frac{p(Z, Y)}{p(Z)p(Y)} \right] \quad (14)$$

Observe that we can rewrite the mutual information as an expected KL divergence by applying the definition of conditional probability.

$$I(Z; Y) = \mathbb{E}_{p(y)} \left[ \text{KL}\left( p(Z|Y) \,||\, p(Z) \right) \right] \quad (15)$$

Since we cannot compute the posterior $p(z|y)$ exactly, we approximate it with the variational distribution $q(z|y)$ (the joint is approximated by $q(z|y)p(y)$ where the latter factor is the data distribution). To the extent that the variational distribution recovers the true posterior, the mutual information can be computed this way. In fact, if we take the learned prior $p(z)$ to be an approximation of the marginal $\int q(z|y)p(y)dy$ it can easily be shown that the thus computed KL term is an upper bound on mutual information (Alemi et al., 2017).

The trouble is that the ELBO (Equation (11)) can be trivially maximised by setting the KL-term to 0 and maximising only the reconstruction term.

---
[2]The term *weak generator* has first been coined by Alemi et al. (2017).

This is especially likely at the beginning of training when the variational approximation does not yet encode much useful information. We can only hope to learn a useful variational distribution if a) the variational approximation is allowed to move away from the prior and b) the resulting increase in the reconstruction term is higher than the increase in the KL-term (i.e. the ELBO increases overall).

Several schemes have been proposed to enable better learning of the variational distribution (Bowman et al., 2016; Kingma et al., 2016; Alemi et al., 2017). Here we use KL scaling and increase the scale gradually until the original objective is recovered. This has the following effect: during the initial learning stage, the KL-term barely contributes to the objective and thus the updates to the variational parameters are driven by the signal from the reconstruction term and hardly restricted by the prior.

Once the scale factor approaches 1 the variational distribution will be highly informative to the generator (assuming sufficiently slow increase of the scale factor). The KL-term can now be minimised by matching the prior to the variational distribution. Notice that up to this point, the prior has hardly been updated. Thus moving the variational approximation back to the prior would likely reduce the reconstruction term since the standard normal prior is not useful for inference purposes. This is in stark contrast to Bowman et al. (2016) whose prior was a *fixed* standard normal distribution. Although they used KL scaling, the KL term could only be decreased by moving the variational approximation back to the fixed prior. This problem disappears in our model where priors are learned.

Moving the prior towards the variational approximation has another desirable effect. The prior can now learn to emulate the variational "look-ahead" mechanism without having access to future contexts itself (recall that the inference model has access to future target tokens). At test time we can thus hope to have learned latent variable distributions that encode information not only about the output at the current position but about future outputs as well.

## 5 Experiments

We report experiments on the IWSLT 2016 data set which contains transcriptions of TED talks and their respective translations. We trained models to translate from English into Arabic, Czech, French and German. The number of sentences for each language after preprocessing is shown in Table 1.

| Data  | Arabic  | Czech   | French  | German  |
|-------|---------|---------|---------|---------|
| Train | 224,125 | 114,389 | 220,399 | 196,883 |
| Dev   | 6,746   | 5,326   | 5,937   | 6,996   |
| Test  | 2,762   | 2,762   | 2,762   | 2,762   |

Table 1: Number of parallel sentence pairs for each language paired with English for IWSLT data.

The vocabulary was split into 50,000 subword units using Google's sentence piece[3] software in its standard settings. As our baseline NMT systems we use Sockeye (Hieber et al., 2017)[4]. Sockeye implements several different NMT models but here we use the standard recurrent attentional model described in Section 2. We report baselines with and without dropout (Srivastava et al., 2014). For dropout a retention probability of 0.5 was used.

As a second baseline we use our own implementation of the model of Zhang et al. (2016) which contains a single sentence-level Gaussian latent variable (SENT). Our implementation differs from theirs in three aspects. First, we feed the last hidden state of the bidirectional encoding into encoding of the source and target sentence into the inference network (Zhang et al. (2016) use the average of all states). Second, the latent variable is smaller in size than the one used by (Zhang et al., 2016).[5] This was done to make their model and the stochastic decoder proposed here as similar as possible. Finally, their implementation was based on groundhog whereas ours builds on Sockeye.

Our stochastic decoder model (SDEC) is also built on top of the basic Sockeye model. It adds the components described in Sections 3 and 4. Recall that the functions that compute the means and standard deviations are implemented by neural nets with a single hidden layer with tanh activation. The width of that layer is twice the size of the latent variable. In our experiments we tested different latent variable sizes and used KL scaling (see Section 4.1). The scale started from 0 and was increased by $1/20{,}000$ after each mini-batch. Thus, at iteration $t$ the scale is $\min(t/20{,}000, 1)$.

All models use 1028 units for the LSTM hid-den state (or 512 for each direction in the bidirectional LSTMs) and 256 for the attention mechansim. Training is done with Adam (Kingma and Ba, 2015). In decoding we use a beam of size 5 and output the most likely word at each position. We deterministically set all latent variables to their mean values during decoding. Monte Carlo decoding (Gal, 2016) is difficult to apply to our setting as it would require sampling entire translations.

---
[3]https://github.com/google/sentencepiece
[4]https://github.com/awslabs/sockeye
[5]We did, however, find that increasing the latent variable size actually hurt performance in our implementation.

**Results** We show the BLEU scores for all models that we tested on the IWSLT data set in Table 2. The stochastic decoder dominates the Sockeye baseline across all 4 languages, and outperforms SENT on most languages. Except on German, there is a trend towards smaller latent variable sizes being more helpful. This is in line with findings by Chung et al. (2015) and Fraccaro et al. (2016) who also used relatively small latent variables. This observation also implies that our model does not improve simply because it has more parameters than the baseline.

That the margin between the SDEC and SENT models is not large was to be expected for two reasons. First, Chung et al. (2015) and Fraccaro et al. (2016) have shown that stochastic RNNs lead to enormous improvements in modelling continuous sequences but only modest increases in performance for discrete sequences (such as natural language). Second, translation performance is measured in BLEU score. We observed that SDEC often reached better ELBO values than SENT indicating a better model fit. How to fully leverage the better modelling ability of stochastic RNNs when producing discrete outputs is a matter of future research.

**Qualitative Analysis** Finally, we would like to demonstrate that our model does indeed capture variation in translation. To this end, we randomly picked sentences from the IWSLT test set and had our model translate them several times, however, the values of the latent variables were sampled instead of fixed. Contrary to the BLEU-based evaluation, beam search was not used in this evaluation in order to avoid interaction between different latent variable samples. See Figure 3 for examples of syntactic and lexical variation. It is important to note that we do not sample from the categorical output distribution. For each target position we pick the most likely word. A non-stochastic NMT system would always yield the same translation in this scenario. Interestingly, when we applied the sampling procedure to the SENT model it did not produce any variation at all, thus behaving like a deterministic NMT system. This supports our initial point that the SENT model is likely insensitive to local variation, a problem that our model was designed to address. Like the model of Bowman et al. (2016), SENT presumably tends to ignore the latent variable.

## 6 Related Work

The stochastic decoder is strongly influenced by previous work on stochastic RNNs. The first such proposal was made by Bayer and Osendorfer (2015) who introduced i.i.d. Gaussian latent variables at each output position. Since their model neglects any sequential dependence of the noise sources, it underperformed on several sequence modeling tasks. Chung et al. (2015) made the latent variables depend on previous information by feeding the previous decoder state into the latent variable sampler. Their inference model did not make use of future elements in the sequence.

Using a "look-ahead" mechanism in the inference net was proposed by Fraccaro et al. (2016) who had a separate stochastic and deterministic RNN layer which both influence the output. Since the stochastic layer in their model depends on the deterministic layer but not vice versa, they could first run the deterministic layer at inference time and then condition the inference net's encoding of the future on the thus obtained features. Like us, they used KL scaling during training.

More recently, Goyal et al. (2017) proposed an auxiliary loss that has the inference net predict future feature representations. This approach yields state-of-the-art results but is still in need of a theoretical justification.

Within translation, Zhang et al. (2016) were the first to incorporate Gaussian variables into an NMT model. Their approach only uses one sentence-level latent variable (corresponding to our $z_0$) and can thus not deal with word-level variation directly. Concurrently to our work, Su et al. (2018) have also proposed a recurrent latent variable model for NMT. Their approach differs from ours in that they do not use a $0^{th}$ latent variable nor a look-ahead mechanism during inference time. Furthermore, their underlying recurrent model is a GRU.

In the wider field of NLP, deep generative mod-els have been applied mostly in monolingual settings such as text generation (Bowman et al., 2016; Semeniuta et al., 2017), morphological analysis (Zhou and Neubig, 2017), dialogue modelling (Wen et al., 2017), question selection (Miao et al., 2016) and summarisation (Miao and Blunsom, 2016).

| Model | Dropout | LatentDim | Arabic | Czech | French | German |
|---|---|---|---|---|---|---|
| Sockeye | None | None | 8.2 | 6.9 | 23.5 | 14.3 |
| Sockeye | 0.5 | None | 8.4 | 7.4 | 24.4 | 15.1 |
| SENT | 0.5 | 64 | 8.4 | 7.3 | 24.8 | 15.3 |
| SENT | 0.5 | 128 | 8.7 | 7.4 | 24.0 | 15.7 |
| SENT | 0.5 | 256 | 8.9 | 7.4 | 24.7 | 15.5 |
| SDEC | 0.5 | 64 | 8.2 | 7.7 | 25.3 | 15.4 |
| SDEC | 0.5 | 128 | 8.8 | 7.5 | 24.2 | 15.6 |
| SDEC | 0.5 | 256 | 8.7 | 7.5 | 23.2 | 15.9 |

Table 2: BLEU scores for different models on the IWSLT data for translation into English. Recall that all SDEC and SENT models used KL scaling during training.

| | |
|---|---|
| Source | Coincidentally, at the same time, the first easy-to-use clinical tests for diagnosing autism were introduced. |
| SENT | Im gleichen Zeitraum wurden die ersten einfachen klinischen Tests für Diagnose getestet. |
| SDEC | Übrigens, zur gleichen Zeit, wurden die ersten einfache klinische Tests für die Diagnose von Autismus eingeführt. |
| SDEC | Übrigens, zur gleichen Zeit, <u>waren</u> die ersten einfache klinische Tests für die Diagnose von Autismus eingeführt <u>worden</u>. |
| Source | They undertook a study of autism prevalence in the general population. |
| SENT | Sie haben eine Studie von Autismus in der allgemeinen Population übernommen. |
| SDEC | Sie entwarfen eine Studie von Autismus in der allgemeinen Bevölkerung. |
| SDEC | Sie <u>führten</u> eine Studie von Autismus in der allgemeinen Population <u>ein</u>. |

Figure 3: Sampled translations from our model (SDEC) and the sentent-level latent variable model (SENT). The first SDEC example shows alternation between the German simple past and past perfect. The past perfect introduces a long range dependency between the main and auxiliary verb (underlined) that the model handles well. The second example shows variation in the lexical realisation of the verb. The second variant uses a particle verb and we again observe a long range dependency between the main verb and its particle (underlined).

## 7 Conclusion and Future Work

We have presented a recurrent decoder for machine translation that uses word-level Gaussian variables to model underlying sources of variation observed in translation corpora. Our experiments confirm our intuition that modelling variation is crucial to the success of machine translation. The proposed model consistently outperforms strong baselines on several language pairs.

As this is the first work that systematically considers word-level variation in NMT, there are lots of research ideas to explore in the future. Here, we list the three which we believe to be most promising.

- Latent factor models: our model only contains one source of variation per word. A latent factor model such as DARN (Gregor et al., 2014) would consider several sources simultaneously. This would also allow us to perform a better analysis of the model behaviour as we could correlate the factors with observed linguistic phenomena.
- Richer prior and variational distributions: The diagonal Gaussian is likely too simple a

distribution to appropriately model the variation in our data. Richer distributions computed by normalising flows (Rezende and Mohamed, 2015; Kingma et al., 2016) will likely improve our model.

- Extension to other architectures: Introducing latent variables into non-autoregressive translation models such as the transformer (Vaswani et al., 2017) should increase their translation ability further.

# 8 Acknowledgements

Philip Schulz and Wilker Aziz were supported by the Dutch Organisation for Scientific Research (NWO) VICI Grant nr. 277-89-002. Trevor Cohn is the recipient of an Australian Research Council Future Fellowship (project number FT130101105).

## A  Gaussian Reparametrisation

Any continuous variable can be transformed into another one by an invertible and differentiable transformation $h$. While we can compute the density of the transformed variable at a given point, we can generally not sample from that density. Fortunately, Gaussian variables can be transformed into standard Gaussian variables. The standard Gaussian distribution is easy to sample from. Let $z \sim \mathcal{N}\left(\mu, \sigma^2\right)$ and recall that in the inference network $\mu$ and $\sigma$ are functions of the variational parameters $\lambda$ which we seek to optimise. We can standardise $z$ by subtracting its mean and dividing by the standard deviation,

$$\epsilon = h(z, \mu, \sigma) = \frac{z - \mu}{\sigma} \ . \tag{16}$$

We write the standard Gaussian density of $\epsilon \sim \mathcal{N}(0, \mathrm{I})$ as $\phi(\epsilon)$.

Differentiating the reconstruction term of the ELBO with respect to the variational parameters $\lambda$ poses a challenge. If we differentiate the expectation we end up with an expression that is itself not an expectation and can thus not be approximated with MC methods (see Equation (11) of the main text). In order to enable MC estimation we replace the measure of the expectation with its transformed version.

$$\begin{aligned} q(z|\mu, \sigma^2) &= \phi(h(z, \mu, \sigma)) \times \left|\frac{d}{dz} h(z, \mu, \sigma)\right| \\ &= \phi(\epsilon) \times \left|\frac{d\epsilon}{dz}\right| \end{aligned} \tag{17}$$

The Jacobian term corrects for the change in volume introduced by the transformation.

Taking the derivative of the ELBO with respect to the variational parameters $\lambda$ is now easy as we can push the derivative operator inside the expectation to obtain stochastic gradient estimates.

$$\frac{\partial}{\partial \lambda} \mathbb{E}_{q(z|\mu,\sigma^2)}[\log p(y|x, z)] = \tag{18a}$$

$$\frac{\partial}{\partial \lambda} \int q(z|\mu, \sigma^2) \log p(y|x, z) \mathrm{d}z = \tag{18b}$$

$$\frac{\partial}{\partial \lambda} \int \phi(\epsilon) \times \left|\frac{d\epsilon}{dz}\right| \log p(y|x, z) \mathrm{d}\epsilon \left|\frac{dz}{d\epsilon}\right| = \tag{18c}$$

$$\int \phi(\epsilon) \times \frac{\partial}{\partial \lambda} \log p(y|x, z) \mathrm{d}\epsilon = \tag{18d}$$

$$\mathbb{E}_{\phi(\epsilon)}\left[\frac{\partial}{\partial \lambda} \log p(y|x, \underbrace{h^{-1}(\epsilon, \mu, \sigma)}_{z})\right] \tag{18e}$$

Notice that the inverse transformation $h^{-1}$ is exactly the one described in Equation (5) of the main text. In line (18c) we have used the substitution rule. As a result the Jacobian terms cancel, leaving an integral whose measure is standard normal.